\documentclass[11pt]{article}
\usepackage[a4paper,margin=1in]{geometry}
\usepackage{graphicx}
\usepackage{hyperref}
\usepackage{amsmath}
\usepackage[super]{cite}

\usepackage[font=small,labelfont=bf,justification=raggedright]{caption}

\title{\textbf{TD-MPC-Opt: Distilling Model-Based \\Multi-Task Reinforcement Learning Agents}}

\author{
  Dmytro Kuzmenko \\
  \small Department of Multimedia Systems, NaUKMA, Kyiv, Ukraine\\
  \small \texttt{kuzmenko@ukma.edu.ua} 
  \and
  Nadiya Shvai \\
  \small Department of Mathematics, NaUKMA, Kyiv, Ukraine\\
  \small \texttt{n.shvay@ukma.edu.ua}
}

\date{\normalsize This is a preprint. The manuscript has been submitted for peer review.}

\begin{document}

\maketitle

\begin{abstract}
We present a novel approach to knowledge transfer in model-based reinforcement learning, addressing the critical challenge of deploying large world models in resource-constrained environments. Our method efficiently distills a high-capacity multi-task agent (317M parameters) into a compact model (1M parameters) on the MT30 benchmark, significantly improving performance across diverse tasks. Our distilled model achieves a state-of-the-art normalized score of 28.45, surpassing the original 1M parameter model score of 18.93. This improvement demonstrates the ability of our distillation technique to capture and consolidate complex multi-task knowledge. We further optimize the distilled model through FP16 post-training quantization, reducing its size by $\sim$50\%. Our approach addresses practical deployment limitations and offers insights into knowledge representation in large world models, paving the way for more efficient and accessible multi-task reinforcement learning systems in robotics and other resource-constrained applications. Code available at \href{https://github.com/dmytro-kuzmenko/td-mpc-opt/}{https://github.com/dmytro-kuzmenko/td-mpc-opt}.
\end{abstract}


% \textbf{AMS Subject Classification:} 68T05, 68T07, 93C41

% Include your sections here
\section{Introduction}

Reinforcement learning (RL) has demonstrated remarkable progress in solving complex tasks across various domains, from playing Atari games \cite{mnih2013playingatarideepreinforcement} to robotic control \cite{levine2016end, andrychowicz2020learning, kalashnikov2018qt}. However, developing agents capable of performing multiple tasks efficiently remains a significant challenge, particularly in resource-constrained environments typical in robotics and real-world applications.

Model-based RL approaches \cite{hafner2020dream, chua2018deep} have emerged as a promising solution, offering improved sample efficiency and generalization capabilities. Recent advancements, such as DreamerV3 \cite{hafner2023mastering} and PWM \cite{georgiev2024pwm} have showcased the potential of large world models in capturing complex dynamics across diverse tasks. However, these approaches often require substantial computational resources, limiting their practical application in resource-constrained settings.

Another development in this field is TD-MPC (Temporal Difference Model Predictive Control) \cite{hansen2022temporal}, which combines elements of RL with model predictive control. This hybrid approach leverages a learned world model for both value estimation and short-horizon planning. The Temporal Difference \cite{sutton1988learning} component allows the agent to learn and improve its model from experience, while the MPC component enables efficient action selection through online planning \cite{garcia2989model}.

TD-MPC2 \cite{hansen2024td}, an extension of TD-MPC, has shown remarkable performance in multi-task continuous control benchmarks. It learns a shared world model across multiple tasks, enabling efficient knowledge transfer and generalization. However, the high-capacity models used in TD-MPC2 (up to 317M parameters) may pose additional challenges for deployment in resource-constrained environments such as robotic systems with limited onboard computing power.

\begin{figure}[htpb]
    \centering
    \includegraphics[width=0.8\columnwidth]{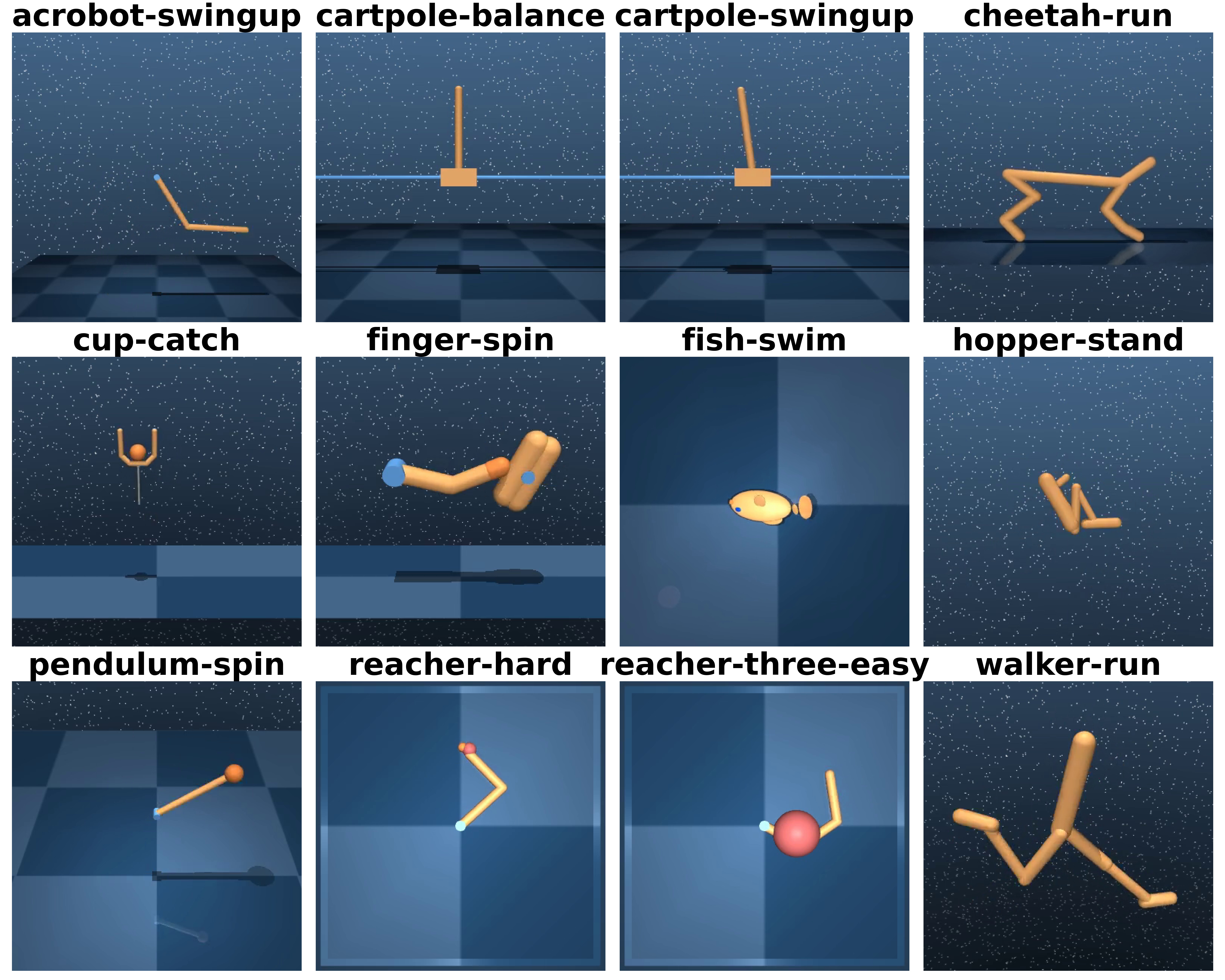}
    \caption{Task visualizations. Visualization of a random initial state for 12 of the 30 tasks that we consider (MT30). The tasks have different action spaces, degrees of freedom, and objectives.}
    \label{fig:mt30_examples}
\end{figure}

This work extends our preliminary results described in \cite{kuzmenko2025knowledgetransfermodelbasedreinforcement} and attempts to address this limitation by exploring knowledge transfer techniques to distill the capabilities of large multi-task agents into more compact and deployment-friendly models. We focus on the MT30 benchmark \cite{hansen2024td}, which comprises tasks from the DM Control \cite{tassa2020dmcontrol}, presenting a diverse set of continuous control challenges with varying observation and action spaces (Figure \ref{fig:mt30_examples}).

The primary contributions of our work are:
\begin{itemize}
\item Development of an efficient reward-based knowledge distillation technique for model-based RL for multi-task offline learning
\item Achievement of state-of-the-art performance (28.45 normalized score) with a compact FP16-quantized 1M parameter model, surpassing the original model by a large margin (+48.5\%) and  improving over a model, trained with the original recipe from scratch (28.12 vs. 27.36, +2.77\%)
\item Demonstration of the effectiveness of longer distillation periods (1M steps) in capturing complex multi-task knowledge
\item Exploration of the impact of \textit{d\_coef} on model performance as well the effect of reduced batch size on efficient knowledge distillation for multi-task scenarios
\item Successful application of FP16 and other quantization types (mixed precision, INT8) to further reduce model size with minimal performance loss
\end{itemize}

Our approach combines traditional teacher-student distillation with quantization techniques, addressing the growing need for lightweight RL models in robotics and real-world applications. 
In the following sections, we detail our methodology, present experimental results on the MT30 benchmark, and discuss the implications of our findings for the future of multi-task model-based RL. 

%%%%%%%%%%%%%%%%%%%%%%%%%%%%%%%%%%%%%%%%%%%%%%%%%%%%%%%%%%%%%%%%%%%%%%%%%%%%%%%%%%%%%%%%%%%%%%%%
\section{Related Work}

RL has progressed immensely in recent years, with advancements in both model-free and model-based approaches. Model-free methods, including Deep Q-Networks (DQN) \cite{mnih2015human}, Policy Gradient techniques \cite{sutton2000policy}, and Soft Actor-Critic (SAC) \cite{haarnoja2018soft}, have demonstrated success in various domains, yet they often struggle with sample efficiency and generalization.

Model-based RL has addressed some of these challenges, offering improved sample efficiency and planning capabilities. Notable approaches include PETS \cite{chua2018deep}, MBPO \cite{janner2019trust}, and DreamerV3 \cite{hafner2023mastering}, which leverage learned world models for decision-making. TD-MPC \cite{hansen2022temporal} and its successor TD-MPC2 \cite{hansen2024td} have further advanced the field by combining model-based RL with model predictive control, showing particular promise in continuous control tasks.

Knowledge distillation, originally introduced by Hinton et al. \cite{hinton2015distilling} for supervised learning, has gained traction in the RL community as a means to transfer knowledge from complex models to simpler ones. In the context of RL, distillation faces unique challenges due to the sequential nature of decision-making and the potential instability of learning processes. Rusu et al. \cite{rusu2015policy} proposed Policy Distillation, adapting the distillation framework to transfer knowledge between deep Q-networks. Their method demonstrated the ability to compress multiple task-specific policies into a single multi-task policy. Building on this, Teh et al. introduced Distral \cite{teh2017distral}, a framework for distilling common knowledge across multiple tasks, improving both multi-task and transfer learning performance.

Parisotto et al. \cite{parisotto2015actor} introduced the Actor-Mimic method, which uses policy distillation to transfer knowledge from multiple task-specific teacher networks to a single multi-task network. Czarnecki et al. \cite{czarnecki2019distilling} proposed an improved policy distillation technique that addresses some limitations of previous methods, such as the need for careful hyperparameter tuning.  In the realm of model-based RL, Kurutach et al. explored model ensemble distillation \cite{kurutach2018model} to improve the stability and performance of model-based policy optimization. Their approach distilled an ensemble of dynamics models into a single model, reducing the negative impact of model bias. More recently, Stooke et al. introduced Decoupled Distillation \cite{stooke2020decoupling}, a technique that separates the distillation of the policy and value function in actor-critic methods, allowing for more efficient knowledge transfer in off-policy RL algorithms.

Multi-task RL has emerged as a crucial area of research, aiming to develop agents capable of performing diverse tasks across different embodiments. Benchmarks such as MT10, MT50 \cite{yu2020meta}, and more recently, the MT30 and MT80 \cite{hansen2024td} benchmarks incorporating tasks from DM Control \cite{tassa2020dmcontrol} and Meta-World \cite{yu2020meta} have been instrumental in evaluating multi-task capabilities of RL agents. These benchmarks present unique challenges, including varying observation and action spaces, which test an agent's ability to generalize across different embodiments and task structures.

Approaches to multi-task RL include shared representation learning \cite{caruana1997multitask}, policy distillation \cite{rusu2015policy}, and model-agnostic meta-learning techniques like MAML \cite{finn2017model}. However, these methods often face challenges such as negative transfer and task interference \cite{zhang2021survey}, highlighting the complexity of balancing task-specific and shared knowledge. Recent work in curriculum learning for RL \cite{narvekar2020curriculum} has shown promise in addressing these challenges by structuring the learning process to gradually increase task complexity.

The increasing complexity of RL models, particularly in multi-task settings, has led to a growing interest in model compression techniques. Knowledge distillation, as discussed earlier, is one approach to this challenge. Additionally, quantization methods, including post-training quantization and quantization-aware training \cite{gholami2021survey}, offer ways to reduce model size while maintaining relative performance. Shin et al. \cite{shin2019deep} explored the use of quantization for deep Q-networks, demonstrating that careful quantization can significantly reduce model size with minimal performance loss. 

Recent trends in RL include the exploration of multi-modal approaches combining vision, language, and control \cite{lee2022multimodal}, and advancements in offline RL \cite{levine2020offline}. These developments are pushing the boundaries of what's possible in RL, but also raise new challenges in terms of computational resources and deployment feasibility. Efficient RL methods, such as Conservative Q-Learning \cite{yu2021conservative}, have shown promise in improving sample efficiency and performance in offline settings, potentially complementing distillation approaches in resource-constrained environments.

Our work builds upon these foundations, particularly in the realm of model-based multi-task RL, by addressing the critical challenge of deploying large world models in resource-constrained environments. By combining knowledge distillation and quantization techniques, we aim to bridge the gap between the impressive capabilities of large models and the practical limitations of real-world deployment across various embodiments and tasks.

%%%%%%%%%%%%%%%%%%%%%%%%%%%%%%%%%%%%%%%%%%%%%%%%%%%%%%%%%%%%%%%%%%%%%%%%%%%%%%%%%%%%%%%%%%%%%%%%
\section{Methods}

Our approach focuses on optimizing large model-based RL agents for efficient multi-task learning through knowledge distillation and model compression. We aim to transfer knowledge from a high-capacity TD-MPC2 model (317M parameters) to a compact 1M parameter model, suitable for deployment in resource-constrained environments. Our method combines traditional teacher-student distillation with quantization techniques, aiming to create new efficient pipelines for lightweight RL models for robotics and real-world applications.

\begin{figure}[htpb]
  \centering
  \includegraphics[width=0.7\linewidth]{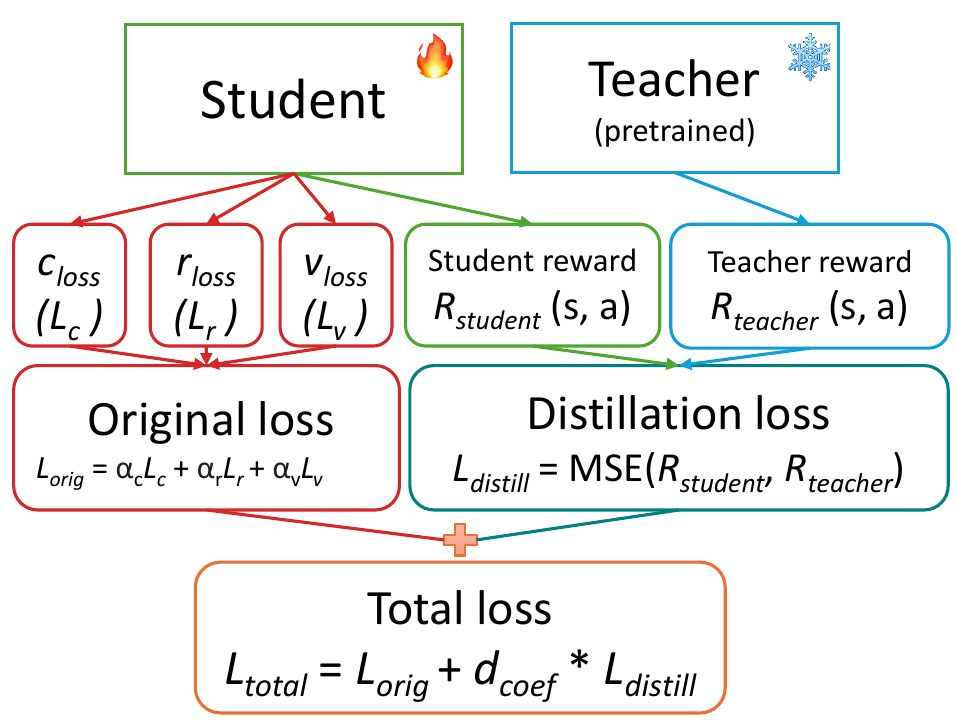}
  \caption{Our distillation approach consists of two main loss function components: the original loss from TD-MPC2 (in red), calculated as a linear combination of consistency, reward, and value losses ($\alpha$ denotes a scaling coefficient for each respective loss); and the distillation loss (in teal) that calculates MSE between student's (green) and teacher's (blue) rewards produced from inferring the same state-action pair. The total loss is a linear combination of both losses, with the distillation loss scaled by the \textit{d\_coef}. In our scenario, the student is trainable while the teacher's weights are frozen.}
  \label{fig:method_fig}
\end{figure}

We employ a teacher-student distillation framework, adapting it to the specific challenges of model-based RL (Figure \ref{fig:method_fig}). This framework consists of two key components: the largest available 317M-parameter TD-MPC2 checkpoint from \cite{hansen2024td} as the teacher; the 1M parameter TD-MPC2 backbone designed to balance performance with deployment efficiency as the student. 

\subsection{Loss Function Adaptation}

We build upon the original TD-MPC2 loss functions (consistency, reward, and value losses) by introducing an additional reward distillation loss. This new loss is computed as the mean squared error (MSE) between the rewards predicted by the teacher and student models:
\[
L_\text{distill} = \text{MSE}(R_\text{teacher}(s, a), R_\text{student}(s, a))
\]
where $R_\text{teacher}$ and $R_\text{student}$ are the reward predictions of the teacher and student models, respectively.

We introduce a distillation coefficient (\textit{d\_coef}) to balance the original TD-MPC2 loss with our new distillation loss:
\[
total\_loss = original\_loss + d_\text{coef} * distillation\_loss
\]
The \textit{d\_coef} acts as a hyperparameter controlling the influence of the teacher model's knowledge on the student model's learning process. We empirically find that values close to 0.5 yield the best results, with 0.4 being optimal for most training setups (Table \ref{tab:big_table}).

\subsection{Evaluation Metric}
We use the normalized score as our primary evaluation metric, consistent with \cite{hansen2024td}. Each task is scored on a scale of 1 to 1000, with the average sum divided by the number of tasks, resulting in a final range of 1-100. This allows for a fair comparison across tasks with different reward scales and difficulties.

\subsection{Training Process}
For training, we take a 317M parameter TD-MPC2 model checkpoint trained on the full MT30 dataset. We then train the 1M parameter student model using our distillation approach. The student is trained on the same dataset but with the additional reward distillation loss. We explore different distillation periods, from 200,000 steps to 1,000,000 steps, to understand the impact of extended distillation on performance, various batch sizes (from 128 to 1024), and multiple values of \textit{d\_coef}. We retain the original architecture and planning hyperparameters adhering to the original idea expressed in TD-MPC2 work\cite{hansen2024td}.

To further optimize our model for real-world deployment, we apply post-training FP16 \cite{micikevicius2018mixed}, mixed precision, and INT8 quantization techniques to our distilled model.

\begin{table}[htpb]
	\caption{Impact of \textit{d\_coef} on student's performance, 200K steps with 317M teacher and batch size of 256. The first two entries indicate attempts to model a next-state latent distillation and the respective mechanism to bridge the latent dimension gap between teacher and student models.}
	% \label{table 1}
        \begin{center}
	\begin{tabular}{cc}
            \hline
		\textit{Distillation coefficient} & \textit{Normalized score} \\
            \hline
            % \hline
            reward + next-state (linear projection) & 7.69 \\
            reward + next-state (PCA) & 8.78 \\
            \hline
            % \hline
		0.05 & 13.61 \\ 
            \hline
            baseline (no distill) & 14.04 \\
            \hline
            0.25 & 14.49 \\ 
            \textbf{0.4} & \textbf{17.85} \\
            0.55 & 16.08 \\ 
            0.6 & 14.83 \\ 
            0.9 & 13.79 \\
            \hline
	\end{tabular}
    \end{center}
    \label{tab:big_table}
\end{table}

\subsection{Experimental Setup}
For the MT30 benchmark, we utilized the full available dataset of 690,000 episodes (345,690,000 transitions). This multi-task benchmark is particularly challenging as it comprises tasks with varying observation and action spaces, requiring the agent to adapt to diverse environments. We conducted our experiments on a desktop PC with a single RTX 3060 GPU (12GB VRAM). 

%%%%%%%%%%%%%%%%%%%%%%%%%%%%%%%%%%%%%%%%%%%%%%%%%%%%%%%%%%%%%%%%%%%%%%%%%%%%%%%%%%%%%%%%%%%%%%%%
\section{Results}

We first reproduced the checkpoint of the 1M parameter model from TD-MPC2  by training from scratch for 200,000 steps with batch size 1024. This served as our primary point of comparison for the distilled models.

\subsection{Short-term Distillation}
We evaluate the sample efficiency of our distillation approach under various training regimes, summarized in Table \ref{tab:sample_efficiency}. For the short-term setting, models were trained for 200,000 steps using different batch sizes. This number of steps was assumed from prior work on MT30 \cite{hansen2024td} and allowed us to reproduce the reported performance of a 1M-parameter model trained from scratch (18.7 normalized score vs 18.93 reported).

Our distilled model achieves a comparable score of 18.11 with the same training duration, though slightly underperforming the from-scratch baseline. However, when using smaller batch sizes (e.g., 256), distillation with an efficient \textit{d\_coef} value provides a clear gain over training from scratch (17.85 vs 14.04). This highlights that the knowledge-distilled agent can reach high performance more efficiently under low-resource constraints. Notably, even at  200K steps and reduced batch size, the distilled policy achieves 17.37 score and approaches full-performance levels with significantly less data exposure.

\begin{table}[htpb]
    \centering
    \caption{Results of different setups of 1M-parameter model distillation vs training from scratch on MT30 benchmark.}
    \vspace{0.5em}  % Adds spacing between caption and top rule
    \begin{tabular}{lccc}
        \hline
        \textit{Method} & \textit{Batch size} & \textit{Training steps, \#} & \textit{Score} \\
        \hline
        distill & 128 & 200K & 17.37 \\
        from scratch & 256 & 200K & 14.04 \\ 
        distill & 256 & 200K & \underline{17.85} \\ 
        \hline
        from scratch & 1024 & 200K & \underline{18.7} \\
        distill & 1024 & 200K & 18.11 \\ 
        \hline
        from scratch & 1024 & 337K & \underline{26.94} \\
        distill & 1024 & 337K & 25.44 \\ 
        \hline
        from scratch & 256 & 1M & 27.36 \\
        distill & 256 & 1M & 28.12 \\ 
        \hline
        \hline
        best FP16 quantized & 256 & 1M & \textbf{28.45} \\
        \hline
    \end{tabular}
    \label{tab:sample_efficiency}
\end{table}

\subsection{Extended Distillation}
We further extended distillation to match the full training length of the original model. For batch size 1024, this translates to approximately 337K steps -- sufficient to cover all 690,000 episodes (or 349 million transitions) in the MT30 benchmark with one batch per step. The aim was to examine whether prolonged distillation allows the student to internalize complex task knowledge more effectively.

While extended distillation closes the performance gap, it still lags behind the from-scratch baseline (25.44 vs 26.94). This suggests that long training alone is not sufficient, and student capacity and alignment with teacher dynamics may impose inherent limits. However, when distillation is extended with a batch size of 256 and 1M steps, the model slightly surpasses the baseline (28.12 vs 27.36), indicating that in some regimes, distillation is not only sample-efficient but can outperform direct training. The highest score (28.45) is achieved using FP16 quantization, demonstrating that sample efficiency and model compression can be jointly optimized.

\begin{figure}[htpb]
    \centering
    \includegraphics[width=0.8\columnwidth]{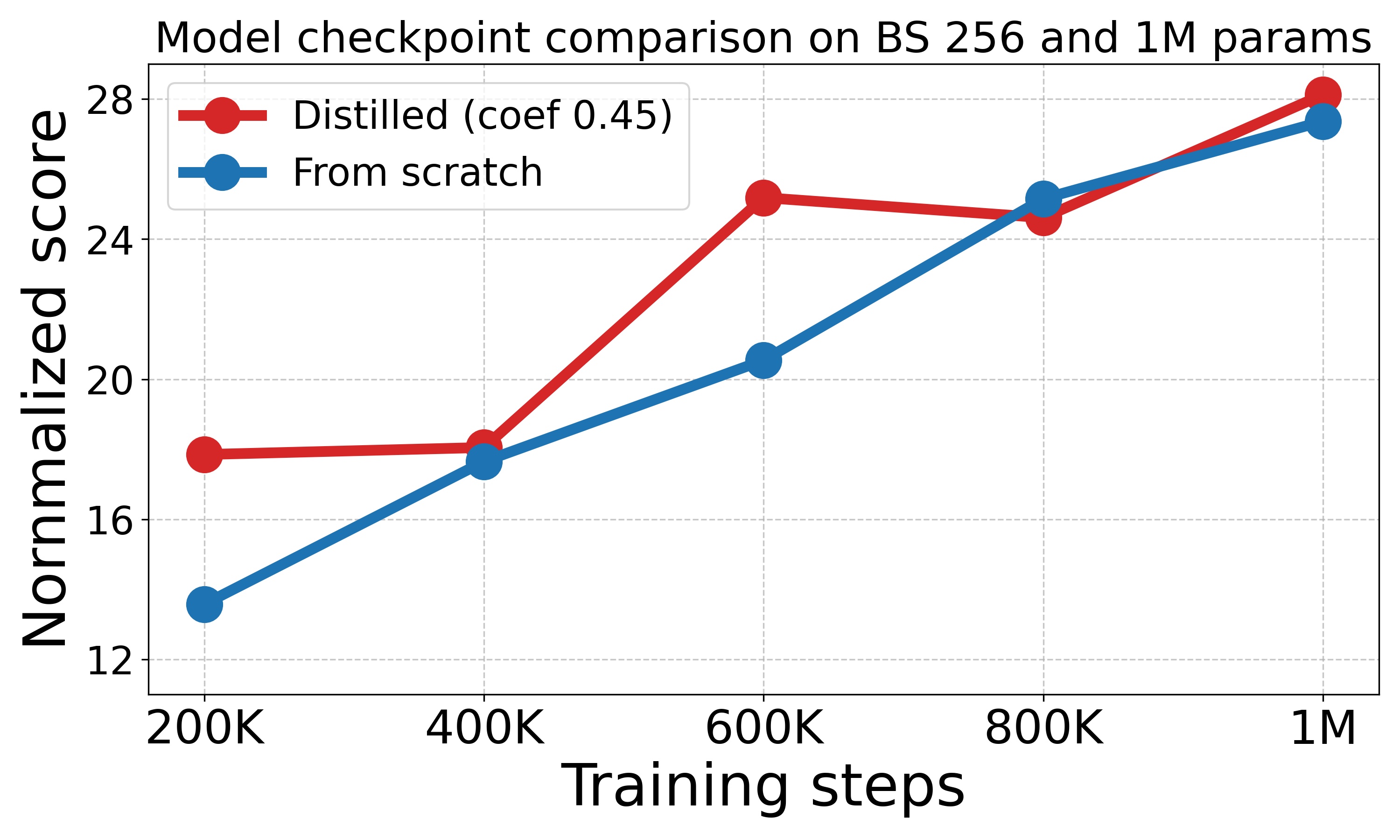}
    \caption{Performance comparison between a knowledge-distilled model (\textit{d\_coef} of 0.45) and a model trained from scratch. Both models have 1M parameters and were trained with a batch size of 256. The normalized scores are shown for five checkpoints, taken at 200K-step intervals up to 1M steps.}
    \label{fig:dist_plot}
\end{figure}

As illustrated in figure \ref{fig:dist_plot}, we compared the performance trajectories of our knowledge-distilled model against a model trained from scratch. Each model with 1M parameters was trained using identical batch sizes of 256. We evaluated their performance at five equidistant checkpoints throughout 1M training steps. The distilled model, utilizing a \textit{d\_coef} of 0.45, demonstrates consistently superior performance, particularly in the early stages of training (200K and 600K steps checkpoints).

\subsection{Batch Size Study}
We investigated how batch size influences the quality of knowledge transfer, convergence speed, and computational efficiency in the distillation of large world models for multi-task RL. Experiments were conducted with batch sizes of 128, 256, and 1024, across both short-term (200K steps) and extended (337K or 1M steps) training regimes (Table \ref{tab:sample_efficiency}). A 128-batch run showed no advantage over 256, highlighting diminishing returns at very small batch sizes. For batch size 1024, we additionally ran a 337K-step setup, so that we cover the entire MT30 dataset and examine the effects of full dataset exposure on distillation quality.

While 1M-step distillation with batch size 256 results in each observation being seen multiple times, we report the outcome as a reference for high-resource settings, leaving deeper analysis for future work. All models were evaluated on the MT30 benchmark using normalized scores for fair cross-task comparison. This design isolates the role of batch size in distillation performance and informs best practices for scalable, resource-aware deployment of distilled policies.

\subsection{Teacher Size Impact}
We compared the performance of student models distilled from two different teacher sizes (Table 1): the second-largest 48M parameter model and the largest 317M parameter model, conceptualizing the relationship between teacher model capacity and distillation effectiveness (Table \ref{tab:teacher_sizes}).

\begin{table}[htpb]
	\caption{Different teacher sizes' effect on distillation. 200K steps, Batch Size 256}
	\label{table_1}
        \begin{center}
	\begin{tabular}{cc} % \toprule
        \hline
	\textit{Teacher config} & \textit{Normalized score} \\ % \midrule
        \hline
        48M & 13.61 \\ 
  	317M & \textbf{17.85} \\ % \bottomrule
        \hline
	\end{tabular}
    \end{center}
    \label{tab:teacher_sizes}
\end{table}

\subsection{Quantization Results}
To further optimize our model for deployment, we applied FP16 quantization to our best-performing distilled model. We then evaluated the quantized model's performance across all tasks to assess the impact of quantization on both model size and task performance.

A mixed precision approach, combining FP16 for sensitive layers and INT8 for others, achieved a 64.6\% size reduction (to 2.8 MiB) but led to a significant performance decrease, with the normalized score dropping to 13.53. This suggests that while mixed precision offers size reduction benefits, it comes at a considerable cost to model performance in the RL context. INT8 (8-bit integer) quantization, a more aggressive compression, reduced the model size by 75.3\% (to 1.95 MiB). However, this came at a substantial cost to performance, with the normalized score plummeting to 5.09. The results are depicted in Table \ref{tab:quant_table}.

\begin{table}[htpb]
    \centering
        \caption{Quantization Results on Distilled Model. 1M steps, Batch Size 256, d\_coef=0.4}
        \begin{center}
        \begin{tabular}{lcccc}
        \hline
        Metric & Original & FP16  & Mixed & INT8 \\
        \hline
        Normalized Score & 28.12 & \textbf{28.45} & 13.53 & 5.09 \\
        Model Size (MiB) & 7.9 & 3.9 & 2.8 & 1.95 \\
        \hline
        \end{tabular}
        \end{center}
    \label{tab:quant_table}
\end{table}

It is important to note that the effects of quantization on RL models may differ from those observed in supervised learning tasks. RL models often have complex dynamics and may be more sensitive to precision loss in certain layers. Therefore, careful evaluation and potentially layer-wise quantization strategies might be necessary for optimal results.

\subsection{Task-specific distillation}
Our experiments reveal that incorporating next-state latent distillation alongside reward distillation presents a significant challenge due to the dimensional mismatch between the teacher (1376-dim) and student (128-dim) models. We attempt to bridge this mismatch with dimensionality reduction techniques, such as linear projection and PCA, which introduce substantial information loss and lead to poor generalization. Linear projection yields a normalized score of 7.69, while PCA marginally improves the score to 8.78, though both underperform the 17.85 achieved with reward-only distillation (Table \ref{tab:big_table}). The additional computational overhead of PCA and its limited gains further underscored its inefficiency.

Our finalized approach, \textbf{reward-only distillation}, proved effective across most tasks, especially in locomotion (\textit{walker-walk}) and manipulation (\textit{reacher-easy}). However, it struggled in tasks requiring long-term planning, sparse rewards, or precise control (e.g. \textit{acrobot-swingup}, \textit{cartpole-swingup-sparse}, \textit{cheetah-jump}) -- all that require next-state prediction.

Latent next-state prediction, a probable necessity for future efficient and robust multi-task agents, did not yield strong results in our experiments due to dimensionality mismatches between teacher and student representations. Further work is needed to address this challenge effectively.

%%%%%%%%%%%%%%%%%%%%%%%%%%%%%%%%%%%%%%%%%%%%%%%%%%%%%%%%%%%%%%%%%%%%%%%%%%%%%%%%%%%%%%%%%%%%%%%%
\section{Discussion}

We demonstrate that knowledge distillation is a viable and sample-efficient strategy for scaling model-based RL to multi-task domains. The strong performance of our distilled agents, especially in extended training settings, confirms that large world models encode rich, transferable representations of task dynamics and reward structures. Crucially, our batch size study revealed that smaller batches (e.g., 256) consistently outperformed larger ones (1024), even in long-horizon distillation. This result challenges the assumption that larger batch sizes are inherently better for sample-efficient knowledge transfer in deep learning.

Our best model, a quantized 1M-parameter student trained for 1M steps with batch size 256, achieves a new state-of-the-art score of 28.45 on the MT30 benchmark (+50.2\% over the previously reported 18.93), using only reward-level distillation. This highlights the high sample efficiency of our method: it compresses a large teacher into a lightweight, high-performing policy that is viable for deployment. Smaller batch sizes appear to enable more frequent model updates and fine-grained learning, particularly beneficial for complex or sparse-reward tasks. With FP16 quantization, we further reduced model size by $\sim$50\%, underscoring the practicality of our approach for real-world applications.

We also examined how batch size and distillation length interact with task complexity, offering insights into when and how smaller models can be trained effectively. These findings inform new training pipelines, where a high-capacity teacher is trained once and distilled into smaller agents tailored for constrained settings, enabling scalable and cost-effective RL deployment.

Despite promising results, our work has limitations. First, we did not validate the distilled models on physical systems. Bridging the gap between simulation and real-world deployment is essential for broader adoption. Second, while our study focused on the MT30 benchmark, evaluating on larger task suites (e.g., MT80) and assessing generalization to unseen tasks remain important directions. Additionally, we have not yet determined the minimum compute budget necessary for effective distillation under various constraints.

Our current approach also relies solely on reward-based distillation. Integrating next-state latent prediction or richer behavioral objectives could be the next step to capturing deeper aspects of the teacher policy, especially for sophisticated control tasks.

In summary, our method enables high-performance, quantized multi-task RL agents with strong sample efficiency. By combining reward distillation with lightweight architectures, we show that compact, deployable agents can retain the competence of large models. This work contributes a scalable and practical pathway toward real-world RL systems suited for diverse and resource-constrained environments.

\bibliographystyle{plain}

\end{document}